\pdfoutput=1

\documentclass[11pt]{article}

\usepackage{acl}

\usepackage{times}
\usepackage{latexsym}

\usepackage[T1]{fontenc}

\usepackage[utf8]{inputenc}

\usepackage{microtype}

\usepackage{inconsolata}

\usepackage{graphicx}
\usepackage{tabularx}
\usepackage{multirow}
\usepackage{hhline}
\usepackage{booktabs}

\newcommand\egg{\textsc{EGG}}
\newcommand\qgsmall{\textsc{\egg-FLAN}}
\newcommand\qglarge{\textsc{\egg-Llama}}

\newcommand{\daggerfootnote}

\title{ Disentangling Questions from 
 Query Generation\\\vspace{0.07cm} for Task-Adaptive Retrieval}

\author{Yoonsang Lee ~~ Minsoo Kim ~~ Seung-won Hwang\textsuperscript{*} \\ Seoul National University \\ \texttt{\{lysianthus, minsoo9574, seungwonh\}@snu.ac.kr}}

\begin{document}
\maketitle
\newcommand\blfootnote[1]{%
  \begingroup
  \renewcommand\thefootnote{}\footnote{\hspace{-0.55\footnotesep}#1}%
  \addtocounter{footnote}{-1}%
  \endgroup
}

\begin{abstract}
This paper studies the problem of information retrieval, to adapt to unseen tasks. Existing work generates synthetic queries from domain-specific documents to jointly train the retriever. However, the conventional query generator assumes the query as a question, thus failing to accommodate general search intents. A more lenient approach incorporates task-adaptive elements, such as few-shot learning with an 137B LLM. In this paper, we challenge a trend equating query and question, and instead conceptualize query generation task as a ``compilation'' of high-level intent into task-adaptive query. Specifically, we propose \egg, a query generator that better adapts to wide search intents expressed in the BeIR benchmark. Our method outperforms baselines and existing models on four tasks with underexplored intents, while utilizing a query generator 47 times smaller than the previous state-of-the-art. Our findings reveal that instructing the LM with explicit search intent is a key aspect of modeling an effective query generator.\footnote{Our code is available at \url{https://github.com/lilys012/metaprompt-QG}.} 
\blfootnote{\textsuperscript{*}Corresponding author.}

\end{abstract}

\section{Introduction}

\begin{figure}[t]
\centerline{\includegraphics[width=0.5\textwidth]{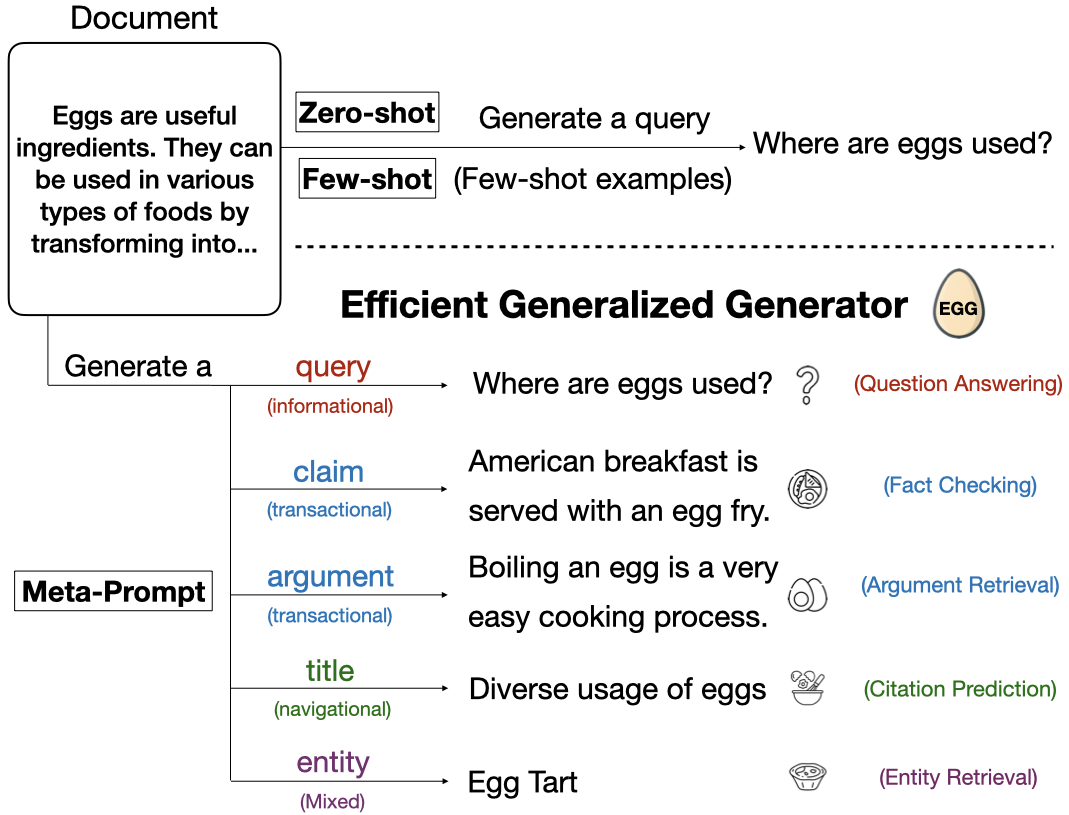}}
\caption{
Overview of our method. Given a document, conventional zero-shot query generator generates questions, while few-shot query generator performs in-context learning with few-shot examples. In contrast, our method reflects diverse search intents utilizing meta-prompts to enhance the generalizability of the query generator. Fact checking can be viewed as {transaction}, where the retriever determines whether the given claim is supported or not, and argument retrieval is similarly so. Citation prediction, presenting a title as the query, represents a {navigational} intent for a specific document, and entity retrieval exhibits a mixture of the intents.
}
\label{fig:overview}
\end{figure}

Information retrieval has significantly facilitated the process of locating relevant documents in response to user requests. With the advent of dense retrieval \citep{Karpukhin2020DensePR}, prior works have concentrated on the supervised alignment of latent spaces within query and passage encoders \citep{Gao2021CondenserAP, Ni2021LargeDE, Santhanam2021ColBERTv2EA}. However, this requires collecting labeled data across numerous domains, which are often unavailable. In such scenarios, where only given the target corpus, existing work focuses on zero-shot query generation to form a synthetic dataset \citep{Cheriton2019FromDT, Ma2021ZeroshotNP}. Representing approaches include GenQ \citep{Thakur2021BEIRAHr}, training a query generator using MSMARCO \citep{Campos2016MSMA}, an extensive question-answering dataset, and Promptagator-Zero \citep{Dai2022PromptagatorFD}, prompting the LLM to generate questions about the documents. While the former inadvertently generates queries in questions forms, being pretrained from MSMARCO queries in questions form, the latter intentionally does so.

We challenge the prevailing trend of equating queries with questions in training.
Though question form may align  with
information intent,  web search intents
are categorized to include more diverse intents~\cite{Broder2002ATO} such as \textit{Informational}, \textit{Navigational}, and \textit{Transactional}. While concentrating on the first category has paid off,  given the prevalence of datasets aligned with informational intent,  generalization is critical, by four non-QA datasets in BeIR~\citep{Thakur2021BEIRAHr} tasks in Figure \ref{fig:overview}.

We thus interpret query generation task as a ``compilation'' of high-level intent into task-adaptive query, to reflect diverse search intents \citep{Bonifacio2022InParsUD, Asai2022TaskawareRW, Hashemi2023DenseRA}. This task has been mostly delegated to in-context learning: Promptagator-Few leverages FLAN 137B \citep{Wei2021FinetunedLM} as a few-shot query generator to better capture the latent intent in query-document pairs. However, challenges may arise in in-context learning \citep{Brown2020LanguageMA} due to small model size \citep{Wei2022EmergentAO}, limited context length \citep{Li2023InContextLW}, or poor quality of examples \citep{Nori2023CanGF}.

To overcome the reliance on expensive few-shot examples and enable the use of small LMs as query generators, we propose \textbf{\egg} (\textbf{E}fficient \textbf{G}eneralized \textbf{G}enerator), which leverages \textit{meta-prompt}\footnote{Following \citet{Nayak2024LearningTG}, we define \textit{meta-prompt} as the prompt that accepts a document and a task attribute as input to generate a task-adaptive query.} to incorporate unique search intents. Our system comes in two model sizes: \qgsmall\:is for scenarios where the model is too small to support in-context learning, for which, we adapt instructions for diverse search intents. However, we find the LM generated queries lack sufficient diversity, thus we ensure the instruction to diversify those. In cases where the model size is just enough to support in-context learning (\qglarge), we first generate \textit{prototype} queries with {meta-prompt} as candidates for in-context query-document pairs. Another key distinction is, we utilize retriever feedback, specifically, relevance ranking. \egg\:not only surpasses both zero-shot and few-shot baselines but also outperforms previous state-of-the-art methods. Our method effectively covers undersupported intents which existing methods often struggle with. 

To summarize, our contributions are as follows: (1) We introduce \egg, which employs meta-prompt and retriever feedback, to integrate unique search intents into query generation. (2) Our method outperforms the baselines, demonstrating its effectiveness. (3) Despite the much smaller size of the query generator, our method achieves the highest overall performance.

\section {Methodology}

\subsection{Task Formulation}
Given the corpus $D^c=\{d_i\}$ and the search intent $e_q$, the goal is to generate queries $\{q^*_i\}$ that correspond well to $e_q$. Equipped with the synthetic pairs $\{d_i, q^*_i\}$, passage\footnote{We use both \textit{document} and \textit{passage} interchangeably.} and query encoders are jointly trained to align their latent embedding spaces. During inference, documents and test queries are each embedded with the trained encoders, and the top-$k$ documents with the highest scores are retrieved. 

In conventional zero-shot methods, $e_q$ is often considered as either the term `query' \citep{Dai2022PromptagatorFD, Wang2023Query2docQE, SaadFalcon2023UDAPDRUD}, or `question' \citep{Sachan2022ImprovingPR, Bonifacio2022InParsUD}, interchangeably. We challenge this assumption by defining $e_q$ as query adaptive to each task, by interpreting $e_q$ as task-specific attribute incorporated it into the meta-prompt.

\subsection{Dataset}
BeIR is a comprehensive benchmark for zero-shot retrieval, including 9 distinct tasks. We focus on tasks that involve underexplored intents and select one dataset from each task. Specifically, we evaluate fact checking task Fever \citep{Thorne2018FEVERAL}, argument retrieval task Arguana \citep{Wachsmuth2018RetrievalOT}, citation prediction task Scidocs \citep{Cohan2020SPECTERDR}, and entity retrieval task DBPedia \citep{Hasibi2017DBpediaEntityVA}. We adopt $e_q$ from the BeIR paper, as presented in Table \ref{tab:style}.

\begin{table}[tbp]
\footnotesize
\centering
\begin{tabular}{c|c|c} \toprule
\multicolumn{1}{c|}{Task} & \multicolumn{1}{c|}{Intent} & \multicolumn{1}{c}{$e_q$} \\ \midrule
Fact Checking & Transactional & Claim \\
Argument Retrieval & Transactional & Argument \\
Citation Prediction & Navigational & Title \\
Entity Retrieval & Mixed & Entity \\
\bottomrule
\end{tabular}
\caption{Query description ($e_q$) of the tasks with underexplored search intents.}
\label{tab:style}
\end{table}

\subsection{Query Generation}

\paragraph{EGG-FLAN} FLAN-T5 \citep{Chung2022ScalingIL} is known for its strong ability to follow instructions \citep{Sun2023EvaluatingTZ}. For each $d_i$, we employ the following meta-prompt to generate $N$ queries $\{q^*_{i_k}\}, k \in (1,N)$ per document: ``\texttt{Write a \{$e_q$\} related to topic of the passage. Do not directly use wordings from the passage. $\{d_i\}$}''. Generated queries can reflect the search intent $e_q$ through the instruction for better alignment of the retrievers. Our pilot study showed that when FLAN-T5 is prompted without specific instruction to write queries in its own words, it tends to extract the same sentence from the document as $N$ queries, leading to worse performance. Therefore, we have added such instruction to encourage the model to attempt paraphrasing and explore various parts of the document.

Our meta-prompt incurs the same computational cost as the zero-shot prompt, which has `query' as the task attribute. Compared to the few-shot method, meta-prompt is significantly shorter than the concatenation of few-shot examples, and even shorter than a single document, highlighting the efficiency aspect.

\paragraph{EGG-LLAMA} 
LLMs can exhibit strong performance via in-context learning, especially when in-context examples are relevant to the given context \citep{Liu2021WhatMG, Lee2023CraftingIE}. However, few-shot methods inevitably employ a fixed set of in-context examples, regardless of the documents. In contrast, we enable in-context learning with \textit{prototype} queries generated by meta-prompts to benefit from relevant documents. Concretely, we first generate {prototype} queries $\{q'_i\}$ that align closely with the search intent of the target task, and then perform in-context learning using these relevant examples $\{P_i\}=(d_i, q'_i)$.

Initially, we generate one $q'_i$ per document with Llama2 model \citep{Touvron2023Llama2O} using the following meta-prompt: ``\texttt{[INST] Read the passage and generate a $\{e_q\}$. [/INST] $\{d_i\}$ $\{e_q\}:$}''.
Subsequently, we perform in-context learning on Llama2 model with $\{P_i\}$ to obtain queries of high relevance and quality to the given document \citep{SaadFalcon2023UDAPDRUD}. With respect to some similarity function $f(i, j)$, we retrieve $M$ relevant examples $P_{i_k}=(d_{i_k}, q'_{i_k}), k \in (1, M)$ for each document. Finally, we generate $\{q^*_{i_k}\}$ with the following template: ``\texttt{Passage: \{$d_{i_1}$\} $\{e_q\}$: \{$q'_{i_1}$\} \ldots  Passage: \{$d_{i_M}$\} $\{e_q\}$: \{$q'_{i_M}$\} Passage: \{$d_i$\} $\{e_q\}$:}'' Since \textit{prototype} queries are tailored to $e_q$, we can expect $\{q^*_{i_k}\}$ to also be aligned with $e_q$.

\subsection{Retriever Training} As we have obtained a task-specific synthetic dataset, we now train a retriever on this dataset. We employ two training frameworks: DPR \citep{Karpukhin2020DensePR} and GPL \citep{Wang2021GPLGP}. DPR maximizes the likelihood of the product between query and positive document embeddings through in-batch negatives, while GPL soft-labels the score of query and positive document using cross-encoder and employs MarginMSE loss~\citep{Hofsttter2020ImprovingEN}.

\section{Experiments}
\subsection{Experimental Setup}
We experiment with FLAN-T5-XL (3B) and Llama2 (7B) models, generating 8 queries per passage and showcasing 4 examples during in-context learning. For \qglarge, we use the dot product between the SimCSE \citep{Gao2021SimCSESC} embeddings of $d_i$ and $d_j$ as $f(i, j)$. We train a DistilBERT TAS-B \citep{Hofsttter2021EfficientlyTA} retriever for each task.

Among unsupervised models, we benchmark against BM25, Contriever \citep{Izacard2021UnsupervisedDI}, and ART \citep{Sachan2022QuestionsAA}. For Generation-based models, we report GenQ, GPL, and two versions of Promptagator. While Pretrain-based models are not our main focus, we also include two representatives GTR \citep{Ni2021LargeDE} and TART \citep{Asai2022TaskawareRW}. These methods train a large retriever on massive pretrain corpus with multiple tasks but do not finetune on the target corpus. We evaluate with nDCG@10 metric, a standard measure for the BeIR benchmark. For further details, please refer to Appendix \ref{sec:appendixA}.

\begin{table}[tbp]
\footnotesize
\begin{center}
\scalebox{0.9}{
\begin{tabular}{ll|cccc|c}
\toprule
&& Fev. & Arg. & Sci. & Dbp. & Avg. \\ \midrule
\multirow{6}{*}{DPR} & FLAN (zero) & 54.4 & 53.9 & 14.8 & 30.4 & 38.4 \\
&FLAN (few) & 38.1 & 22.9 & 16.7 & 31.5 & 27.3 \\
&\qgsmall & \textbf{69.5} & \textbf{60.1} & \textbf{18.6} & \textbf{33.6} & \textbf{45.5} \\ \cmidrule(lr){2-7}
&Llama2 (zero) & 60.9 & 59.0 & 16.0 & \textbf{34.2} & 42.5 \\
&Llama2 (few) & 67.2 & \textbf{62.0} & 17.5 & 30.2 & 44.2 \\
&\qglarge & \textbf{67.6} & 61.2 & \textbf{18.2} & 32.5 & \textbf{44.9} \\ \midrule
\multirow{6}{*}{GPL}&FLAN (zero) & 73.2 & 55.8 & 14.9 & 39.6 & 45.9 \\
&FLAN (few) & 77.0 & 3.5 & 16.3 & 37.7 & 33.6 \\
&\qgsmall & \textbf{79.4} & \textbf{58.7} & \textbf{16.9} & \textbf{40.0} & \textbf{48.8} \\ \cmidrule(lr){2-7}
&Llama2 (zero) & 73.0 & 56.4 & 15.0 & \textbf{40.2} & 46.2 \\
&Llama2 (few) & \textbf{79.8} & 55.8 & \textbf{17.0} & 39.1 & 47.9 \\
&\qglarge & 78.3 & \textbf{57.1} & \textbf{17.0} & \textbf{40.2} & \textbf{48.2} \\
\bottomrule
\end{tabular}
}
\caption{Results of \egg, zero-shot, and few-shot baselines with two training methods, DPR and GPL. \egg\:consistently outperforms the baselines. We bold the highest score for each experiment. Fev, Arg, Sci, Dbp stand for Fever, Arguana, Scidocs, and DBPedia.}
\label{tab:baseline}
\end{center}
\end{table}

\begin{table*}[tbp]
\footnotesize
\centering
\begin{tabular}{ll|cc|cccc|c} \toprule
&& QG & retriever & Fever & Arguana & Scidocs & DBPedia & Avg. \\ \midrule
\multirow{3}{*}{\textit{Unsupervised}}&BM25 & - & -  & 75.3 & 31.5 & 15.8 & 31.3 & 38.5 \\
&Contriever & - & 110M & 68.2 & 37.9 & 14.9 & 29.2 & 37.6 \\ 
&ART & - & 220M & 72.4 & 32.2 & 14.4 & 36.3 & 38.8 \\ \midrule
\multirow{2}{*}{\textit{Pretrain-based}}&GTR-XXL & - & 4.8B & 74.0 & 54.0 & 16.1 & 40.8 & 46.2 \\
& TART & - & 1.5B & -$^{**}$ & 51.5 & \underline{18.7} & \underline{46.8} & - \\ \midrule
\multirow{8}{*}{\textit{Generation-based}}&GenQ & 220M\textsuperscript{*} & 66M & 66.9 & 49.3 & 14.3 & 32.8 & 40.8 \\
&GPL & 220M\textsuperscript{*} & 66M & 75.9 & 55.7 & 16.9 & 38.4 & 46.7 \\ 
&Promptagator-Zero & 137B & 110M & 76.2 & 53.8 & 16.3 & 36.4 & 45.7 \\
&Promptagator-Few & 137B & 110M & 77.0 & 59.4 & 18.5 & 38.0 & 48.2 \\ \cmidrule{2-9}
&\multicolumn{1}{l|}{DPR + \qgsmall} & 3B & 66M & 69.5 & 60.1 & \textbf{18.6} & 33.6 & 45.5 \\
&\multicolumn{1}{l|}{\:\:\:\:\:\:\:\:\:\:+ \qglarge} & 7B & 66M & 67.6  & \textbf{\underline{61.2}} & 18.2 & 32.5 & 44.9 \\
&\multicolumn{1}{l|}{GPL + \qgsmall} & 3B & 66M & \underline{\textbf{79.4}} & 58.7 & 16.9 & 40.0 & \textbf{\underline{48.8}} \\
&\multicolumn{1}{l|}{\:\:\:\:\:\:\:\:\:\:+ \qglarge} & 7B & 66M & 78.3 & 57.1 & 17.0 & \textbf{40.2} & 48.2 \\ 
\bottomrule
\end{tabular}
\caption{Model performances across four BeIR tasks in nDCG@10. QG and retriever indicates the model size of query generator and retriever. DPR+\qgsmall\:represents the retriever trained on \qgsmall-generated queries with DPR. Bold and underline indicate the best score among {Generation-based} models and all models. (*) GenQ and GPL further finetune the generator on MSMARCO. (**) Fever is included in the train corpus of TART.} 
\label{tab:results}
\end{table*}

\subsection{Baselines}
Considering different pipelines in existing works, we establish zero-shot and few-shot baselines for a controlled study. Following the conventional zero-shot assumption, we define $e_q$ as the term `query' to generate questions. For few-shot baseline, we employ 4-shot examples for Llama2 and 1-shot example for FLAN-T5, given its limited context size. We follow the same template used in in-context learning for \qglarge.

\section{Results}
\subsection{Main Results}
We compare the baselines and our method in Table \ref{tab:baseline}. Both \qgsmall\:and \qglarge\:exhibit the highest performance among the baselines, underscoring the benefits of incorporating search intents through meta-prompts. Moreover, \qgsmall\:demonstrates a significant gain over its few-shot baseline, as small LMs may struggle to handle few-shot examples. These results indicate that the queries generated by our method effectively cater to various search intents.

While Llama2 exhibits superior baseline performance than FLAN-T5, \qgsmall\: outperforms \qglarge, despite its smaller size and lower costs. Additionally, GPL training method shows a better average score than DPR, primarily due to the gains in Fever and DBPedia. Nevertheless, \egg\:with DPR exhibits marginal improvements in the other two tasks.

\subsection{Overall Performance}
Table \ref{tab:results} describes the overall performance. \qgsmall\:with GPL achieves the top rank overall, while \qglarge\:with GPL attains the second, tied with Promptagator-Few. Compared to {Pretrain-based} models, our method performs better on Arguana and Fever, and comparably on Scidocs, despite much smaller retriever size and fewer pretrain tasks. As {Generation-based} methods can adjust to unseen tasks using synthetic queries, our work suggests a promising avenue for developing a more versatile query generator.
\subsection{Analysis}

\paragraph{Ablation Study}
\label{lab:abl}
We conduct an ablation study on the effect of in-context learning in \qglarge. We train the retriever with 8 \textit{prototype} queries without performing in-context learning. As illustrated in Table \ref{tab:ablation}, in-context learning enhances performance, especially when the generated queries are longer and in a form of sentences (claim, argument vs. title, entity).

\begin{table}[tbp]
\footnotesize
\begin{center}
\begin{tabular}{cccc}
\toprule
Fever & Arguana & Scidocs & DBPedia \\ \midrule
61.9 \textcolor{red}{(-5.7)} & 60.3 \textcolor{red}{(-0.9)} & 18.0 \textcolor{red}{(-0.2)} & 31.9 \textcolor{red}{(-0.6)} \\
\bottomrule
\end{tabular}
\caption{Results of retrievers trained with DPR on 8 prototype queries. Performance drops across all datasets compared to DPR+\qglarge. We put the difference in parenthesis and colour it red.}
\label{tab:ablation}
\end{center}
\end{table}

\renewcommand{\arraystretch}{1.2}
\begin{table*}[htbp]
\footnotesize
\centering 
\begin{tabular}{c|c|p{11.5cm}} \toprule
\multicolumn{1}{c|}{\textbf{Dataset}} & \multicolumn{2}{c}{\textbf{Generated query}} \\ \midrule 
Fever & Zero-shot & Who is the author of the novel ``Don Quixote''? \\
& Few-shot & Sancho Panza's novel Don Quixote written by Don Miguel de Cervantes Saavedra. \\
& {\egg} & Sancho Panza is a humorous and insightful character in the novel Don Quixote who provides commentary on the events of the story and offers a unique perspective on the world. \\
& Gold & Sancho Panza is a fictional character in a novel written by a Spanish writer born in the 17th century. \\ \midrule
Arguana & Zero-shot & What are the ethical and environmental impacts of meat production and consumption, and how do they compare to plant-based food systems? \\
& Few-shot & The meat industry is not green While many vegetarians and vegans may think
that their diet is environmentally friendly, the reality is that the meat industry is
not green. Meat production is a major contributor to greenhouse gas emissions,
deforestation, and water pollution. \\
& {\egg} & Growing crops and vegetation can also be bad for the environment due to the use of fertilisers and pesticides and the destruction of forests and wildlife habitats, as well as the inefficiency of agricultural land use in some countries. \\
& Gold & Being vegetarian helps the environment  Becoming a vegetarian is an environmentally friendly thing to do. Modern farming is one of the main sources of pollution in our rivers. Beef farming is one of the main causes of deforestation, and as long as people continue to buy fast food in their billions, there will be a financial incentive to continue cutting down trees to make room for cattle. (...) \\
\bottomrule
\end{tabular}
\caption{Examples of generated queries with different methods. We truncate too long queries due to space limit. We observe that the queries generated with our method demonstrate high similarity to gold queries, with respect to $e_q$.}
\label{tab:genqueries_abl_main}
\end{table*}

\paragraph{Qualitative Analysis}
In this section, we compare \egg-generated (\qglarge) queries to other queries. As shown in Table \ref{tab:genqueries_abl_main}\footnote{We report full results in Table \ref{tab:genqueries_abl} in the appendix.}, we observe that zero-shot queries are in a question form, which is {semantically} far apart from gold queries. On the other hand, \egg-generated and few-shot queries demonstrate a similar form {and content} to the gold queries. This example illustrates the efficacy of our method on par with few-shot learning, which has already demonstrated its strong performance on various tasks.
Meanwhile, since Arguana retrieves {counter arguments}, we find \egg-generated queries semantically contradict the gold query. Exploring more specific intents may enable constructing of better synthetic queries.

\section {Related Work}
In scenarios where only the target corpus is available \citep{Izacard2021UnsupervisedDI, Ni2021LargeDE, Gao2022PreciseZD}, existing works create synthetic labels by generating queries from documents. A common approach is to train a query generator on large QA datasets \citep{Cheriton2019FromDT, Ma2021ZeroshotNP}.

Recent work prompts LLMs to generate synthetic queries from documents. There has been active research on training {neural rerankers} with synthetic queries \citep{Sachan2022ImprovingPR, Chandradevan2024DUQGenEU, Huang2024InstUPRI}. InPars~\citep{Bonifacio2022InParsUD} performs few-shot learning, while InPars-Light \citep{Boytsov2023InParsLightCU} extends InPars by employing cost efficient recipes. UDAPDR~\citep{SaadFalcon2023UDAPDRUD} generates a small number of queries using GPT-3 and iteratively generates a large number of queries with cheaper model to train rerankers. \citet{Almeida2024ExploringEZ} leverages small LMs to generate questions starting with common initiators, such as What, How, and When. Meanwhile, training task-adaptive retriever is underexplored, only being studied by Promptagator \citep{Dai2022PromptagatorFD}, which leverages few-shot examples to capture the latent intents, albeit at much higher computational costs. 

Another line of study provides specific instructions along with the queries and builds a general-purpose retriever \citep{Asai2022TaskawareRW, Oh2024INSTRUCTIRAB, Weller2024FollowIREA}. This involves large sizes of LLMs during inference, while our method establishes small, task-specific retrievers.

\section{Conclusion}
In this work, we present \egg, a novel approach designed to overcome the shortcomings of previous query generation methods. We propose two designs for \egg, distinguished by their model sizes, to improve the diversity and quality of synthetic queries while effectively capturing the search intent of the task using meta-prompts. Our approach demonstrates superior performance across four tasks, suggesting a promising direction for query generation involving underexplored search intents.

\section*{Limitations}
Some tasks exhibit mixed search intents, such as DBPedia and NFCorpus \citep{Boteva2016AFL} datasets. While we have adopted the most representative description of the query, provided by the authors of BeIR, considering multiple candidates for $e_q$ may augment the performance. Moreover, our study focuses on the commonly affordable sizes of LMs. We reserve the exploration of other variants of FLAN-T5 and Llama2 models to future investigations, seeking to discern their capacity to specialize in certain tasks. Lastly, utilizing a reranker instead of a retriever has demonstrated high performance \citep{Dai2022PromptagatorFD, Bonifacio2022InParsUD, SaadFalcon2023UDAPDRUD}, despite its expensive computations. The performance of our method could be further enhanced with the incorporation of the reranker.

\section*{Acknowledgment}
This work was partly supported by Korea Institute for Advancement of Technology (KIAT) grant funded by the Korea Government (Ministry of Education) (P0025681-G02P22450002201-10054408, Semiconductor-Specialized University), 
and Institute of Information \& communications Technology Planning \& Evaluation (IITP) grant funded by the Korea government (MSIT) [NO.2021-0-0268, Artificial Intelligence Innovation Hub (Artificial Intelligence Institute, Seoul National University)]

\bibliography{anthology,custom}
\clearpage

\appendix

\section{Experimental Details}
\label{sec:appendixA}
We utilize the publicly available FLAN-T5-XL and Llama2 checkpoints\footnote{https://huggingface.co/google/flan-t5-xl, https://huggingface.co/meta-llama/Llama-2-7b-hf, https://huggingface.co/meta-llama/Llama-2-7b-chat-hf}. When generating prototype queries, since we provide instructions, we leverage the chat variant of Llama2 as it is known to be finetuned on instructions. When performing in-context learning, we employ the base model.
In instances where passages exceed 350 tokens, they are truncated, and query sampling is executed with a temperature of 1.0, employing parameters \textit{k = 25} and \textit{p = 0.95}. 
We randomly sample 100K documents if the corpus size exceeds. 
For training the DistilBERT-TASB retriever, a batch size of 75 is adopted. If the corpus size is larger than 60K, a single epoch is conducted; otherwise, 3 epochs are performed. 
The training process incorporates a learning rate of 2e-5 and a warming step of 1000. 
Generating queries with \qgsmall\:are conducted on a single RTX 3090 GPU and generating queries with \qglarge\: are conducted on 4 RTX A6000 GPUs. Query generation with \qgsmall\:took 15 hours in total and \qglarge\: took 75 hours in total. Training with DPR took maximum 1-2 hours per each dataset. We did not modify any training pipeline of GPL.\footnote{https://github.com/UKPLab/gpl}

\section{Query Examples}
We provide examples of 3 generated queries for each dataset. Table \ref{tab:genqueries_flan} and \ref{tab:genqueries_llama} presents queries generated with \qgsmall and \qglarge, respectively.

\begin{table*}[htbp]
\footnotesize
\centering 
\begin{tabular}{c|c|p{11.5cm}} \toprule
\multicolumn{1}{c|}{\textbf{Dataset}} & \multicolumn{2}{c}{\textbf{Generated query}} \\ \midrule 
Fever & Zero-shot & Who is the author of the novel ``Don Quixote''? \\ 
& Few-shot & Sancho Panza's novel Don Quixote written by Don Miguel de Cervantes Saavedra. \\
& \egg\:w/o ICL & ``Sancho Panza's character provides a commentary on the societal issues of the time, particularly those related to class and social hierarchy. \\ 
& {\egg} & Sancho Panza is a humorous and insightful character in the novel Don Quixote who provides commentary on the events of the story and offers a unique perspective on the world. \\ 
& Gold & Sancho Panza is a fictional character in a novel written by a Spanish writer born in the 17th century. \\ \midrule
Arguana & Zero-shot & What are the ethical and environmental impacts of meat production and consumption, and how do they compare to plant-based food systems? \\ 
& Few-shot & The meat industry is not green  While many vegetarians and vegans may think that their diet is environmentally friendly, the reality is that the meat industry is not green.  Meat production is a major contributor to greenhouse gas emissions, deforestation, and water pollution. \\
& \egg\:w/o ICL & ``Do vegetarians really help the environment?'', Simon Farrell offers an opposing view, arguing that vegetarianism may not be the most effective way to reduce the environmental impact of food production. According to Farrell, livestock farming creates unique environments, such as chalk down land and mountain \\ 
& {\egg} & Growing crops and vegetation can also be bad for the environment due to the use of fertilisers and pesticides and the destruction of forests and wildlife habitats, as well as the inefficiency of agricultural land use in some countries. \\ 
& Gold & Being vegetarian helps the environment  Becoming a vegetarian is an environmentally friendly thing to do. Modern farming is one of the main sources of pollution in our rivers. Beef farming is one of the main causes of deforestation, and as long as people continue to buy fast food in their billions, there will be a financial incentive to continue cutting down trees to make room for cattle. (...) \\ \midrule
Scidocs & Zero-shot & What are the key differences between RMSProp and ESGD adaptive learning rate schemes for non-convex optimization? \\ 
& Few-shot & Escape from Saddle Points: A Survey of Non-Convex Optimization Techniques for Deep Learning \\
& \egg\:w/o ICL & Equilibration Preconditioner and Adaptive Learning Rate Schemes for Non-Convex Optimization \\ 
& {\egg} & Equilibration-Based Learning Rates for Deep Neural Networks with Non-Convex Losses \\ 
& Gold & Train longer, generalize better: closing the generalization gap in large batch training of neural networks \\ \midrule
DBPedia & Zero-shot & What is the population density of Steele Township based on the 2010 census? \\  
& Few-shot & township Rowan county North Carolina United States \\
& \egg\:w/o ICL & Steele Township \\ 
& {\egg} & Steele Township, Rowan County, North Carolina \\
& Gold & rowan university \\ 
\bottomrule
\end{tabular}
\caption{The table is formatted the same as Table \ref{tab:genqueries_abl_main}. \egg\:w/o ICL indicates \qglarge\:without in-context learning stage, while \egg\:indicates the full \qglarge.}
\label{tab:genqueries_abl}
\end{table*}

\newpage

\begin{table*}[htbp]
\footnotesize
\centering 
\begin{tabular}{c|p{13cm}} \toprule
\multicolumn{1}{c|}{\textbf{Dataset}} & \multicolumn{1}{c}{\textbf{Generated query}} \\ \midrule 
Fever & Monochamus adamitus is a species of beetle in the Cerambycidae family. \\ 
& Der Klassiker is the name given in German to the match between two German football clubs. \\ 
& Bootstrapping populations for parametric inference. \\ \midrule
Arguana & Animals are sentient beings who can feel pleasure and pain. Animal suffering is just as serious as human suffering. Therefore it is immoral to kill animals for food when we do not need to do so. \\ 
& Sport and politics are separate and should be kept separate \\ 
& The Heathrow Airport has been at capacity since it was built and will continue to be. \\ \midrule
Scidocs & WhatsApp Usage Patterns and Prediction Models \\ 
& Random Walk with Restart on Large Graphs Using Block Elimination\\ 
& Context Suggestion for User-Oriented Recommender Systems \\ \midrule
DBPedia & Jindo Island \\ 
& Game of Thrones (season 5) \\ 
& 2002 IIHF World Junior Ice Hockey Championships \\ 
\bottomrule
\end{tabular}
\caption{Examples of generated queries with \qgsmall. 3 examples are displayed per each dataset.}
\label{tab:genqueries_flan}
\end{table*}

\begin{table*}[htbp]
\footnotesize
\centering 
\begin{tabular}{c|p{13cm}} \toprule
\multicolumn{1}{c|}{\textbf{Dataset}} & \multicolumn{1}{c}{\textbf{Generated query}} \\ \midrule
Fever & Monochamus adamitus is threatened by habitat loss and fragmentation, which is caused by logging and agricultural activities. \\ 
& The matches between Bayern Munich and Borussia Dortmund are considered to be some of the biggest and most exciting in the German football leagues. \\ 
& A bootstrap is a technique for approximating an unknown population distribution from a known sample drawn from it. \\ \midrule
Arguana & Killing animals for food is unjustified and unnecessary, and can be replaced with plant-based or lab-grown alternatives that do not require the killing of animals. \\ 
& The Euro 2012 football tournament should not be used for political posturing and grandstanding. \\ 
& Heathrow airport must expand in order to maintain its competitiveness and avoid falling behind other European airports.\\ \midrule
Scidocs & A Study of WhatsApp Messaging and Behavior: Predictive Models and User Characteristics \\ 
& Fast and Accurate Random Walks with Restarts on Large Graphs Using Block Elimination \\ 
& Context Suggestion: User-Oriented Context Recommendation in Recommender Systems \\ \midrule
DBPedia & Battle of Myeongnyang \\ 
& Game of Thrones Season 5 \\ 
& 2002 Men's World Ice Hockey Championships \\
\bottomrule
\end{tabular}
\caption{Examples of generated queries with \qglarge. 3 examples are displayed per each dataset.}
\label{tab:genqueries_llama}
\end{table*}

\end{document}